\documentclass{article}

\usepackage{arxiv}

\usepackage[utf8]{inputenc} 
\usepackage[T1]{fontenc}    
\usepackage{hyperref}       
\usepackage{url}            
\usepackage{booktabs}       
\usepackage{amsfonts}       
\usepackage{nicefrac}       
\usepackage{microtype}      
\usepackage{lipsum}
\usepackage{graphicx}
\usepackage{multirow}
\usepackage{float}
\usepackage{amsmath}
\usepackage{comment}

\title{A Detailed Comparative Study of Open Source Deep Learning Frameworks}

\author{
  Ghadeer Al-Bdour\\
  Faculty of Computer and Information Technology\\
  Jordan University of Science and Technology\\
  Irbid, Jordan \\
  \texttt{gwalbdour13@cit.just.edu.jo} \\
   \And
 Raffi Al-Qurran \\
  Faculty of Computer and Information Technology\\
  Jordan University of Science and Technology\\
  Irbid, Jordan \\
  \texttt{rlalqurran11@cit.just.edu.jo} \\
  \AND
  Mahmoud Al-Ayyoub\\
  Faculty of Computer and Information Technology\\
  Jordan University of Science and Technology \\
  Irbid, Jordan \\
  \texttt{maalshbool@just.edu.jo} \\
  \And
  Ali Shatnawi \\
  Faculty of Computer and Information Technology\\
  Jordan University of Science and Technology \\
  Irbid, Jordan \\
  \texttt{ali@just.edu.jo} \\
}

\begin{document}
\maketitle

\begin{abstract}
Deep Learning (DL) is one of the hottest fields in machine learning as DL approaches produced results superior to the state-of-the-art in several areas such as image processing and natural language processing (NLP). To foster the growth of DL, several open source frameworks appeared providing implementations of the most common DL algorithms. These frameworks vary in the algorithms they support and in the quality of their implementations. The purpose of this work is to provide a qualitative and quantitative comparison among three of the most popular and most comprehensive DL frameworks (namely Google's TensorFlow, University of Montreal's Theano and Microsoft's CNTK). The ultimate goal of this work is to help end users make an informed decision about the best DL framework that suits their needs and resources. To ensure that our study is as comprehensive as possible, we conduct several experiments using multiple benchmark datasets from different fields (image processing, NLP, etc.) and measure the performance of the frameworks' implementations of different DL algorithms. For most of our experiments, we find out that CNTK's implementations are superior to the other ones under consideration.
\end{abstract}


\section{Introduction}
\label{sec:intro}

Deep Learning (DL) is the hottest field in Machine Learning (ML). The idea of DL is to train a multi-layer Neural Network (NN) on a dataset in order to allow it to handle real world tasks. Although the theoretical concepts behind DL are not new, DL has enjoyed a surge of interest over the past decade due to many factors including its successful application to several problems (many of which have great commercial potentials) and the improved affordability of the 
required computing infrastructure.

DL approaches have significantly outperformed state-of-the-art approaches in many classical problems of many fields such as image processing, computer vision, speech processing, natural language processing (NLP), etc. Moreover, the scientific community (from both the academia and the industry) has quickly and massively adopted DL. Open source implementations of successful DL algorithms quickly appeared on code sharing websites, 
and were subsequently used by many researchers in different fields.

Several DL frameworks exist such as TensorFlow, Theano, CNTK, Caffe, Torch, Neon, pylearn, etc. Each one of these frameworks has different features and performance characteristics. Further, each framework utilizes different techniques to optimize its implementation of DL algorithms. Although the same algorithm is implemented in different frameworks, the performance of the different implementations can vary greatly. Researchers/practitioners looking to employ such algorithms in their research or application face a difficult choice, since the number of different implementations is high and the effort invested by the research community in scientifically comparing these implementations is limited.

In this work, we aim at providing qualitative and quantitative comparison between popular open source DL frameworks. To be more specific, we focus on three very popular DL frameworks, namely Theano (from MILA Lab, University of Montreal), TensorFlow (from Google), and CNTK (from Microsoft). These frameworks support multi-core CPUs as well as multiple GPUs. All of them import cuDNN, which is a DL library from NVIDIA that supports
highly tuned implementations for standard routines such as forward and backward convolution, normalization, pooling, and activation layers.\footnote{The frameworks now have updates which were not present during the writing of this paper.}
We compare these frameworks by training different NN architectures on five different standard benchmark datasets for various tasks in image processing, computer vision and NLP.\footnote{In an earlier version of this work~\cite{shatnawi2018comparative}, we only considered two datasets.}
Despite their importance, comparative studies like ours that focus on performance issues are rare.
A comparative study of the frameworks is important in order to enable people who are interested in applying DL in their research and/or applications to make informed decisions about which of the existing frameworks suits their needs.

The rest of this paper is organized as follows.
Sections~\ref{sec:related} presents our survey of the literature, while Section~\ref{sec:nn} explains in details what NN are. Section~\ref{sec:frameworks} discusses the frameworks, the way they were used to train the datasets and a brief comparison between them.
The methodology we follow is discussed in Section~\ref{sec:method}.
Experimental results and the discussion are detailed in Section~\ref{sec:res}.
The work is concluded with final thoughts presented in Section~\ref{sec:conc}.

\section{Literature Survey}
\label{sec:related}
Only few researchers did a comparative study between state-of-the-art DL frameworks running on different hardware platforms (CPU and GPU) to highlight the advantages and limitations for each framework when applied on different deep NN architectures, and to enable developers to optimize the running performance of DL frameworks.

One of the first and most prominent examples of comparative studies between DL frameworks was carried out by Bahrampour et al.~\cite{bahrampour2015comparative}. The authors compared five DL frameworks: Caffe, Neon, TensorFlow, Theano, and Torch, in terms of speed (gradient computation time and forward time), hardware utilization and extensibility (ability to support different types of DL architectures) after applying various convolutional algorithms on the aforementioned frameworks. They conducted their experiments on a single machine for both CPU (multithreaded) and GPU (NVIDIA Titan X) environments.
The comparison between frameworks was carried out by training convolutional and stacked autoencoder (AE) networks on the MNIST and ImageNet datasets. They also trained long short-term memory (LSTM) networks~\cite{hochreiter1997long} on the IMDB dataset~\cite{maas2011learning}.\footnote{More details will be given later about these neural network architectures and these datasets.}
The authors reported several observations/findings.
In terms of extensibility, they found that Theano and Torch were the best as they can support various DL architectures and libraries. Moreover, they found that TensorFlow was a very flexible framework especially when used in different parts of the computational graph. Finally, emphasizing ease of use, they noticed that Caffe was the easiest.
In terms of performance, they noticed that Torch was the best
for training and testing their DL architectures on a CPU platform. Theano came in second while Neon gave the worst performance.
On a GPU platform, for convolutional and fully connected networks, they found that Torch was the best followed by Theano. Moreover, they noticed that Theano was the fastest on small networks and Torch was the fastest on large networks followed by Neon. For recurrent networks (LSTM), they found that Theano's results were the best in terms of performance. TensorFlow on single GPU was the worst compared to other studied frameworks.

Shi et al.~\cite{shi2016benchmarking} did a comparative study between several DL frameworks including Caffe, MXNet, CNTK, TensorFlow, and Torch. They considered three types of NN including; fully connected NN (FCN), convolutional NN (CNN) and recurrent NN (RNN). Moreover, they used different hardware environments including two CPU platforms and three GPU platforms.
They considered the running time and the convergence rate as the metrics to evaluate the selected frameworks. They used synthetic datasets to measure running time performance and real-world datasets to measure the convergence rate in their experiments.
The results were as follows.
In synthetic datasets, they evaluated the performance of FCN using a large NN (FCN-s). They used AlexNet and ResNet-50 on ImageNet dataset.
For real-world datasets, they applied MNIST dataset using a small FCN (FCN-R). Moreover, they applied CIFAR10 dataset using AlexNet-R and ResNet-56. For RNN, they chose two LSTM layers for testing. 
After experimentation, they found that all tested frameworks achieved significant speed-up using GPU over CPU.
For CPU platform, they found that TensorFlow was the best compared to other tools. On a single GPU, Caffe, CNTK and Torch performed better than MXNet and TensorFlow on FCN implementations. For small CNN, Caffe and CNTK achieved good performances. For RNN (LSTM), CNTK was the fastest as it was five to ten times better than the other tools. Finally, on multi-GPU platforms, all implementations had higher throughput and convergence rate.

Goldsborough~\cite{goldsborough2016tour} showed the timeline of ML software libraries for DL. He focused on TensorFlow's results and its basic properties including computational paradigms, its distributed model and programming interface. He compared TensorFlow to other DL frameworks including Theano, Torch and Caffe, qualitatively and quantitatively.
In qualitative terms, he compared aforementioned frameworks using several categories including frontends, programming model style, how gradients are computed, and distributing the execution of computational graph. Table~\ref{table:peter} shows a summary of this comparison.
In quantitative terms, he reviewed works (such as~\cite{bahrampour2015comparative,al2016theano}), which contain comparisons between TensorFlow and other DL frameworks.
From LeNet benchmark in~\cite{samuel1959some}, he noted that TensorFlow was ranked second after Torch in forward and backward measures, but, in terms of performance, TensorFlow came at the last rank compared to tested frameworks. From the results on~\cite{hinton2006reducing}'s benchmark for convolutional network models, he noted that TensorFlow came at the second place behind Torch in forward and backward propagation time. Finally, in~\cite{martinez2013learning}, the benchmarks were CNN models including AlexNet architecture, and an LSTM network operating on the Penn TreeBank dataset~\cite{marcus1993building}. He noted that TensorFlow was the best framework for small model followed by Theano then Torch. For large models, TensorFlow came at the second rank after Theano, and Torch came at the last place.

\begin{table}
\centering
\caption{Goldsborough~\cite{goldsborough2016tour}'s qualitative comparison of DL frameworks}\label{table:peter}
\begin{tabular}{ccccc}
\hline
Library & Frontends & Style & Gradients & Distributed Execution\\
\hline
TensorFlow & Python,C++ & Declarative & Symbolic & \checkmark\\
Theano & Python & Declarative & Symbolic & X\\
Torch & LuaJIT & Imperative & Explicit & X\\
Caffe & Protobuf & Imperative & Explicit & X\\
\hline
\end{tabular}
\end{table}

Chintala\footnote{\url{https://github.com/soumith/convnet-benchmarks}}
applied different ImageNet benchmarks for a variety convolutional network types including AlexNet, GoogleNet, Overfeat, and OxfordNet using different open source DL frameworks such as Caffe, Theano, Torch, TensorFlow, Chainer, etc.
He conducted his experiments on NVIDIA Titan X GPU and two new packages for Fast Fourier Transform (FFT) computation~\cite{vasilache2014fast}. The first one is based on the NVIDIA library (cuFFT) and another on a Facebook FFT (fbfft). Moreover, he used native version of each DL frameworks. After experimentation, he found that using fbfft resulted in a speed-up over cuFFT for all applied CNN. Moreover, he found that the fastest framework for CNN was Torch followed by Tensorflow.

Theano development team of Al-Rfou et al.~\cite{al2016theano} discussed the Theano framework, its features, how to use it, and showed recent improvements on it. They did a performance comparison between Theano and other frameworks including Torch and TensorFlow on three types of ML models including CNN, RNN and sequence-to-sequence mapping RNN. Finally, they showed the computation speed using multiple GPUs.
The results were as follows.
On a single GPU platform, for CNN, they found the processing time for four different convolutional models (AlexNet~\cite{krizhevsky2014one}, OverFeat~\cite{sermanet2013overfeat}, VGG~\cite{simonyan2014very}, and GoogLeNet~\cite{szegedy2015going}) on ImageNet dataset. They reported results for each framework per minibatch for forward and backward pass. They found that Theano was slower than Torch and TensorFlow. However, in overall performance-wise, they were close to each other.
For RNN, they used LSTM on Penn TreeBank dataset~\cite{marcus1993building} and reported the results on small, medium and large LSTM models. They found that Theano was in the second place after TensorFlow for the small model, but Theano was the fastest for the medium and large models. They also showed that Torch was slower than Theano on all tested models.
Finally, for sequence-to-sequence model~\cite{yao2015describing}, the input was video frames and the output was the English sentence describing the input. The input video frame was preprocessed by a GoogLeNet (pre-trained for classification on ImageNet). They compared Theano to TensorFlow and excluded Torch because there was no available implementation in Torch. They reported the processing time per minibatch using three different batch sizes (32, 64 and 128). They found that Theano was the fastest on smaller batches. However, on large ones, TensorFlow was the superior one. They repeated the previous RNN model (LSTM) on multi-GPU platforms (2-GPUs and 4-GPUs) using platoon. Their measured processing speed when synchronizing after each batch was found to provide speed-ups between 1.6X and 1.7X for 2-GPUs and 3.2X for 4-GPUs.
When synchronizing after every 100 batches, they found a 2X speed-up for 2-GPUs and 3.9X-4X speed-up for 4-GPUs.

Kovalev et al.~\cite{kovalev2016deep} presented a comparative study between five DL frameworks: Theano, Torch, Caffe, TensorFlow, and DeepLearning4J, in terms of training and prediction speed and classification accuracy. They used MNIST dataset of handwritten digits for testing five FCN frameworks. Their computation experiments were applied only on CPU; they reported out the results for two kinds of scaling factors applied on FCN networks including changing the network's \textit{depth} (number of internal layers) and changing the network's \textit{width} (number of neurons). Moreover, they tested the NN with two different activation functions, Tanh and ReLU.
In Tanh nonlinearity function, they found that the training time is approximately 30 seconds when they changed the number of layers from one to four in all frameworks except for DeepLearning4J. For DeepLearning4J, the training time started from 140 seconds for one layer and grew up to 210 seconds when they used four layers. For the prediction time, they found that Theano, Torch, Caffe, and TensorFlow consumed less than 0.4 second. However, for DeepLearning4J, it started from 0.75 second when using one layer and increased up to 1.1 second for four layers. In terms of classification accuracy, they found that Theano, DeepLearning4J, and Caffe achieved high accuracy starting from 94\% for one layer and going up to 98\% for four layers. For Torch and TensorFlow, the accuracy dropped with increasing the number of network layers.
With ReLU nonlinearity function, the training time was much lower compared with the case of Tanh function. As for the prediction time, they observed that the use of ReLU gave results that were similar to Tanh results. In terms of accuracy, they found that Torch's accuracy grew up while increasing the number of layers, whereas other frameworks had the same behavior as with Tanh.
Finally, they changed the number of neurons in the hidden layers of networks for ReLU function only, and reported the speed and accuracy values. They found that DeepLearning4J framework was the slowest in training and prediction times, and time consumed increased with the increase in the number of neurons. The final classification accuracy when changing the internal layer sizes from 64 to 1024 neurons remained around 97\% for Theano, Caffe, TensorFlow, and DeepLearning4J. However, in the case of Torch, the classification accuracy grew with the growing layer size (started from 70\% and reached 98\%).

Bastien et al.\cite{bastien2012theano} suggested a new feature to be added to the main features of Theano in order to improve its performance in different benchmarks.
They conducted a comparative study (in terms of features and performance) between Theano and Torch7 on NN benchmarks, and between Theano and RNNLM on RNN benchmarks. In their comparison, they used three learning architectures: \textit{logistic regression}, \textit{NN with one hidden layer} (500 units) and \textit{Deep NN (DNN) with three hidden layers} (1000 hidden units each).
They found that when applying NN on CPU using one hidden layer models without using mini-batches, Theano's results overcome Torch7. However, on the logistic regression benchmark, Torch7 thrived because, at each call, it decreased the amount of performed computation. On the GPU, with batch size equal to one, Torch7 overcame Theano. When using mini-batches, Theano was faster than (or has an equivalent speed to) Torch7 on all three learning architectures under consideration. When they applied RNN on Theano and RNNLM with batch size of one, they found that RNNLM was faster than Theano on smaller models. However,  for bigger sizes, Theano was faster.

Ding et al.~\cite{ding2014theano} introduced an implementation of Theano-based AlexNet on ImageNet dataset and applied it on multiple GPUs in order to accelerate the training process. They compared their results of Theano to Caffe library which runs on single GPU in terms of training time. On a single GPU with batch size equal to 256, Caffe was shown to be faster than Theano. However, on 2-GPUs with batch size equal to 128, they found that Theano was faster than Caffe on 1-GPU.

Dai and Berleant~\cite{dai2019benchmarking} studied benchmarking principles and machine learning hardware setup including GPUs, FPGAs and ASICs. Moreover, they studied deep learning software frameworks. They introduced 11 qualitative benchmarking metrics for hardware devices and six metrics for deep learning software frameworks. Moreover, they compared 18 deep learning frameworks and divided them into three categories namely \textit{mature frameworks} (including Caffe, Facebook Caffe2, Chainer, DyNet, MXNet, CNTK, Tensorflow, Keras, Neon, PlaidML, Pytorch and Theano), \textit{developing frameworks} (including Apache SINGA, BigDL, DL4J and PaddlePaddle) and \textit{inactive frameworks} (including Torch and Purine). The 11 qualitative benchmarking hardware aspects for GPUs, FPGAs and ASICs discussed computing performance, low latency, energy efficiency, compatibility, research costs, research risks, upgradability, scalability, chip price, ubicomp and time to market. On the other hand, the six qualitative benchmarking metrics for DL frameworks discussed the licence type, interface codes (API), compatible hardware, reliability, tested DL networks and tested datasets that encompass wide range of datasets such as image datasets, voice datasets and text datasets as well. They also mentioned the MLPerf benchmarking organization that offers useful benchmarks to evaluate training and inference on DL hardware devices. MLPerf benchmarks include benchmark metrics and datasets such as ImageNet and COCO image datasets, WMT English-German translation datasets and MovieLens-20M recommendation datasets. Another evaluation criteria mentioned was DL algorithms and DL frameworks such Tensorflow, Pytorch, MXNet, Caffe and Sinan.

Coleman et al.~\cite{coleman2017dawnbench} introduced DAWNBench a benchmark focused in end-to-end training time to achieve a SOTA accuracy level, as well as inference time accompanied with that accuracy. They studied how different optimizations, including choice of optimizer, stochastic depth and multi-GPU training affect end-to-end training performance. The initial release of DAWNBench provided end-to-end and inference tasks such as image classification on CIFAR-10 and ImageNet as well as question answering on SQuaD and reference implementations for each task. DAWNBench differs from other benchmarking platforms (Baidu DeepBench, Fathom, and Tensorflow Benchmark) because it focuses on end-to-end performance, where the others uses time needed to train on a single minibatch of data as key metric, while disregrading the resulting accuracy of the trained model. Moreover, other benchmarks focus on timimg individual low-level operations utilized in DL computations, while DAWNBench measures time to a pre-specified level of accuracy taking into account both hardware and statistical performance. Training procedure used for batch size 128, an SGD with a weight decay of 0.0005 and momentum of 0.9, learning rate of 0.01 for five epochs used as well, then proceed with an initial learning rate of 0.1 for 90 epochs. the learning rate is decayed by a factor of 10 every 45 epochs and the training terminated after 185 epochs. Regarding augmentation process, the initial learning rate was linearly scaled using base learning rate of 0.1 corresponding to a batch size of 128. The authors considered three optimization techniques. \textit{Adam} is an adaptive optimizer for gradient descent that reports considerable speed-ups over other adaptive optimizers when training CNN on CIFAR-10. \textit{Single-node multi-GPU training}, using 4 GPUs in two different settings, first with the same minibatch size (128) as that used in the baseline approach, but distributed across the 4 GPUs. Second, with effectively a minibatch size multiplied by four, where every GPU is given a minibatch size of 128. \textit{Stochastic Depth} is another optimizer used that can be thought of as a form of regularization similar to dropout. It reports improvements in training and time accuracy.  


In previous studies, the comparison goal focused only on processing time. None of those comparative studies dealt with CPU and GPU utilization or memory consumption. This work covered these metrics to find which of the considered frameworks achieve the best performance.
Finally and most importantly, the comparisons involved more datasets from more fields compared with previous studies.

\section{Neural Networks}
\label{sec:nn}

We start this section with a glimpse of
history
related to neural networks. 

\subsection{Single Layer Neural Network}
\begin{figure}
    \centering
    \includegraphics[width=0.3\textwidth]{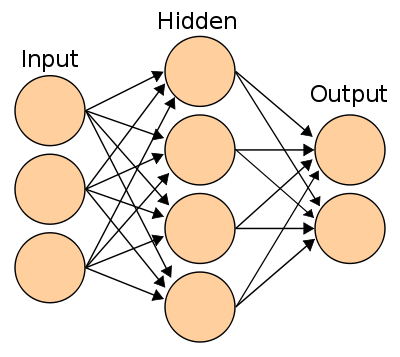}
    \caption[Single layer neural network]{Single layer neural network.\protect\footnotemark}
    \label{fig:fig8}
\end{figure}

A single layer NN is a network that consists of a single hidden layer between the input layer and the output layer. The hidden layer has many units called neurons. See Figure~\ref{fig:fig8}.

\footnotetext{\url{https://en.wikibooks.org/wiki/Artificial_Neural_Networks/Neural_Network_Basics}}

\subsubsection{Artificial Neurons}
A neuron is the building block of the human brain. Neurons process and transmit signals in the form of electrical and chemical pulses. These signals are passed through the neurons via synapses. The same theory applies to artificial neurons where these units when combined with each other perform as a single impenetrable network which takes input signals and performs complex calculations to produce output. This process is depicted in Equation~\ref{eq:an}, where $w_i$ represents connection weights and $x_i$ represents the input values. The left side of the equation $Y$ represents the outputs.
\begin{equation} 
Y = f (\sum_{i=0}^{n} w_ix_i)
\label{eq:an}
\end{equation}

\subsubsection{The Perceptron}
\begin{figure}
    \centering
    \includegraphics[width=0.6\textwidth]{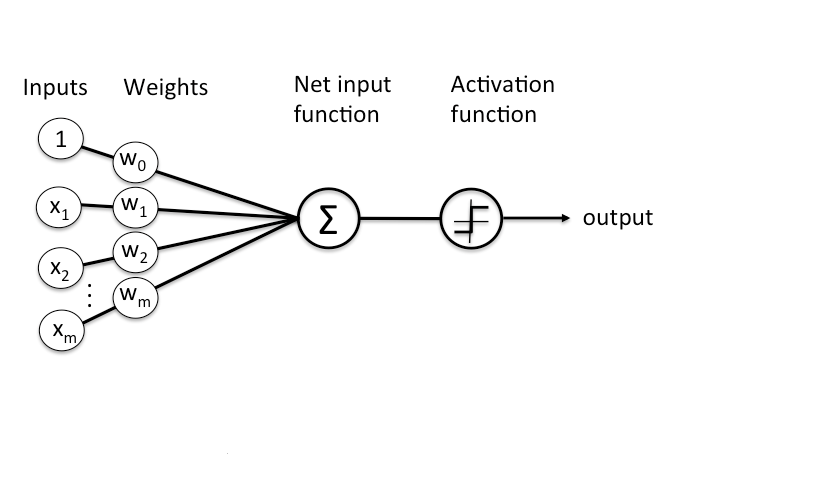}
    \vspace{-0.5in}
    \caption[Rosenblatt's perceptron]{Rosenblatt's perceptron.\protect\footnotemark}
    \label{fig:fig10}
\end{figure}

The perceptron algorithm was devised in 1957 by Rosenblatt~\cite{rosenblatt1957perceptron} for image recognition tasks. A perceptron is the basic form of a NN as it does not contain any hidden layer. Moreover, the input signals are fed directly to output neurons using a series of weights as shown in Figure~\ref{fig:fig10}. The perceptron contains only an input layer and one output layer that consists of one or more artificial neurons.

\footnotetext{\url{https://goo.gl/h9cAJm}}

\begin{figure}
    \centering
    \includegraphics[width=0.5\textwidth]{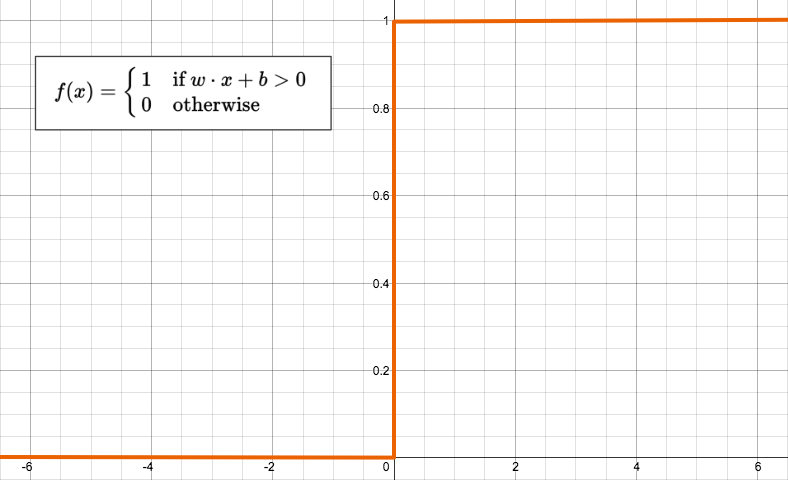}
    \caption[Perceptron's activation function]{Perceptron's activation function.\protect\footnotemark}
    \label{fig:fig11}
\end{figure}

A perceptron is considered a binary classifier as it has only two possible results: 0 or 1. This result is determined by computing a single output from multiple input values. This is done by computing a weighted sum of input values and then putting the output through some nonlinear activation function like the Heaviside step function (as Threshold Function)~\cite{debnath2014integral}, as shown in Figure~\ref{fig:fig11}. The output neuron in the output layer connects to all inputs to produce one output value.

\footnotetext {\url{https://appliedgo.net/perceptron/}}

The advantages of the perceptron include
simplicity of its architecture and its ``light'' computation requirement allowing its efficient use with very large datasets. The main drawback of the perceptron is that it only learns linearly-separable functions. In order to solve this problem, a multilayer perceptron (also known as DNN) was suggested by Ivakhnenko et al.~\cite{ivakhnenko1966cybernetic} to get more powerful learning mechanisms.

\subsubsection{Activation Function}
\begin{figure}
    \centering
    \includegraphics [width=0.5\textwidth]{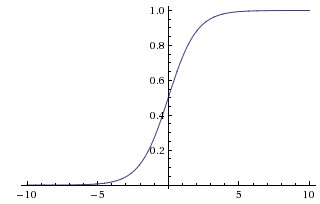}
    \caption[Sigmoid function]{Sigmoid function.\protect\footnotemark}
    \label{fig:fig3}
\end{figure}

An activation function is a nonlinear function that is used in different types of NN. It takes weighted input data and transforms them into a nonlinear output by performing some mathematical operations on them such as matrix multiplication between inputs and weights. Regarding DL implementations, nonlinear activation functions create complex features with every layer. Implementations with a linear activation function would behave like a single-layer network (no matter how many hidden layers) because summing these layers would give just another linear function. This is the reason why nonlinear activation functions are used more widely at DL networks. However, it is possible that some NN may contain neurons with linear activation function in the output layer. These neurons require a nonlinear activation function in previous parts of the network. There are several types of activation functions including the sigmoid function shown in Figure~\ref{fig:fig3}.

\footnotetext{ \url{http://cs231n.github.io/neural-networks-1/}}

The sigmoid function has the following mathematical form.
\begin{equation}
\sigma(x)=\frac{1}{1+e^{-x}}
\end{equation}
It takes a real-value number and converts it into values in the range $[0, 1]$ (large negative numbers are converted to 0 and large positive numbers are converted to 1). Its main drawbacks are the vanishing gradient problem and the fact that its output is not zero-centered.

\begin{figure}
    \centering
    \includegraphics[width=0.5\textwidth]{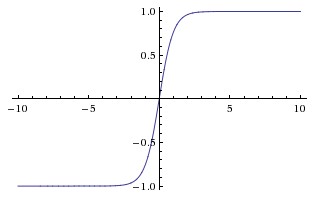}
    \caption[Tanh function]{Tanh function.\protect\footnotemark[8]}
    \label{fig:fig4}
\end{figure}

Another activation function, called \textit{Tanh}, is shown in Figure~\ref{fig:fig4} and it has the following mathematical form.
\begin{equation}
\text{Tanh}(x)=2\sigma (2x)-1
\end{equation}
It is a nonlinear activation function that takes a real-value number and converts it into the range $[-1, 1]$. It causes the vanishing gradient like sigmoid, but its output is zero-centered which enables the Tanh nonlinearity to be used more widely than the sigmoid function.

\begin{figure}
    \centering
    \includegraphics[width=0.5\textwidth]{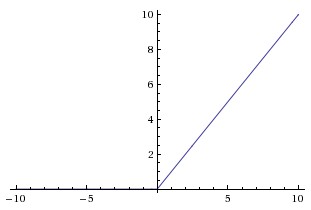}
    \caption[ReLU function]{ReLU function.\protect\footnotemark[8]}
    \label{fig:fig5}
\end{figure}

The \textit{Rectified Linear Unit (ReLU)} activation function (Figure~\ref{fig:fig5}) has the following formula.
\begin{equation}
\text{RelU}(x)=\max(0,x)
\end{equation}
It was found to be more accelerated than the Tanh and sigmoid functions due to its linear and non-saturating form. Moreover, it can be implemented in a less expensive way compared with Tanh and sigmoid. Unfortunately, a ReLU unit may ``die'' during training where it outputs zero value for any given input. This happens when the input to its units are negative or after a large negative bias value is inputted for its weights (gradients will be zero). ReLUs cannot recover from this problem because they will not modify the weights (block backpropagation).

The \textit{Maxout}~\cite{goodfellow2013maxout} function
is a nonlinear activation function that applies dot product between the weights and data, and its output is the maximum of a set of inputs. The Maxout neuron uses the following formula.
\begin{equation}
\text{Maxout}(x)=\max(w_{1}^{T}x+b_1,w_{2}^{T}x+b2)
\end{equation}
It is to be noted that the ReLU function is a special case of the Maxout function. The Maxout neuron has all the benefits of a ReLU unit and does not have its drawbacks (dying unit), which makes Maxout one of the most common activation functions used in DL networks.

Another form of activation functions is called the Logistic Regression. It is a regression model developed in 1958 by Cox~\cite{cox1958regression},
that estimates the relationship between statistical input variables to make prediction of an output variable. It uses the logistic sigmoid function to generate a prediction as to which of multiple classes the input data belongs. Logistic regression is used for different areas like medical and social sciences where it is used for analytical purposes and interpretation of results from experiments. It is used for very large datasets because of its simplicity and speed. The final layer at a DL algorithm can be constructed using a logistic regression, where the network has multiple feature learning layers that pass features into a logistic regression layer to classify inputs.

Logistic regression can be binomial, ordinal or multinomial. In the binomial type, the observed output for a dependent variable has two possibilities: 0 or 1. In multinomial logistic regression, the output can have more than two types (e.g., ``Disease A'' vs ``Disease B'' vs ``Disease C'') which are not ordered. In ordinal type, the dependent variables must be ordered.

In linear regression algorithms (using least square), Gradient Descent (GD) is an optimization algorithm that is used to find the values of parameters in a way that minimizes the cost function and the least square error. If a huge dataset is trained using GD, calculating the parameters will be expensive and take long time. If there are millions of sample points, in every iteration, GD must go through these points to calculate the parameters. To solve these problems, a variation of GD called Stochastic GD (SGD) is used. SGD is an optimization method used to train models including support vector machines (SVM), logistic regression, graphical models, etc.
To calculate the parameters in SGD, a sample of training set or one training value is used instead of using the entire sample in every iteration. This method is much faster and less costly than GD.

Backpropagation of errors is a learning method that is used to train NN. It is used along with an optimization algorithm such as GD. The modern version of backpropagation was proposed by Linnainmaa~\cite{linnainmaa1970representation},
where he published the general method for automatic differentiation (AD) of discrete connected networks of nested differentiable functions. Backpropagation repeatedly performs two steps for training the network; propagation and weights update. At the first step, the input vector is forward propagated layer-by-layer until it reaches the output layer and produces the output of this vector. For each of the neurons in the output layer, an error value is calculated using a loss function, which compares the output of the network to the desired output. It then calculates the gradient of the loss function with respect to the weights. Backpropagation sends these error values backwards starting from the output layer until it reaches the first layer. At the second step, this gradient value is fed to an optimization method (e.g., GD) to update the weights in order to minimize the loss function.
Backpropagation is considered as a supervised learning method, because it requires the knowledge of the desired output to calculate the loss function gradient, but some unsupervised networks such as AE can use it.

The rapid improvement in DL methods makes the training of any DL network a complex and time consuming process. To address these issues, many software tools (frameworks) have appeared to develop these methods in an easy and efficient manner.

\subsection{Multilayer Neural Networks}
\begin{figure}
    \centering
    \includegraphics[width=0.5\textwidth]{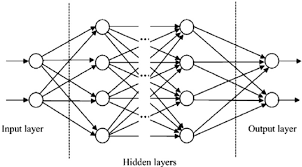}
    \caption[Multilayer Neural Network ]{Multilayer Neural Network.\protect\footnotemark}
    \label{fig:fig9}
\end{figure}

A multilayer NN is a network consisting of more than one hidden layer between the input layer and the output layer, as shown in Figure~\ref{fig:fig9}.
One popular example is the Multilayer Perceptron (MLP) network, which consists of an input layer, one or more hidden layers of computation neurons and an output layer. MLP can learn linear and nonlinear functions in contrast to the single layer perceptron that only supports linear functions. MLP has a large number of features. Moreover, it uses backpropagation technique for training the network. Each node in its hidden layers is a neuron that applies a nonlinear activation function. The input values are passed from the input nodes to the first hidden layer, which applies some calculations to them using the activation function. The resulting signals are then passed as input signals to the next hidden layer. This
procedure is repeated until the signals reach the output layer.

\footnotetext{\url{https://bit.ly/2KgyOdY}}

\subsection{Deep Learning (DL)}

The wide adoption of Deep Neural Networks (DNN) gave rise to the new field of \textit{Deep Learning} (DL). The DNN is simply a NN with more than one hidden layer of nonlinear processing units which extract the features by passing input data from a layer to another until a desirable output is produced. One of the most used algorithms in DL is the Backpropagation algorithm, which is used to train the network by updating the weights of the connections between sequential hidden layers. This process is repeated many times until the output matches the desired output.

There are several types of DL architectures such as DNN, Convolutional DNN (CDNN), Deep Belief Networks (DBN) and RNN. Approaches based on these architectures have been achieving significant performance improvements when applied to several tasks in speech recognition, computer vision, NLP, etc.

\begin{figure}
    \centering
    \includegraphics[width=0.5\textwidth]{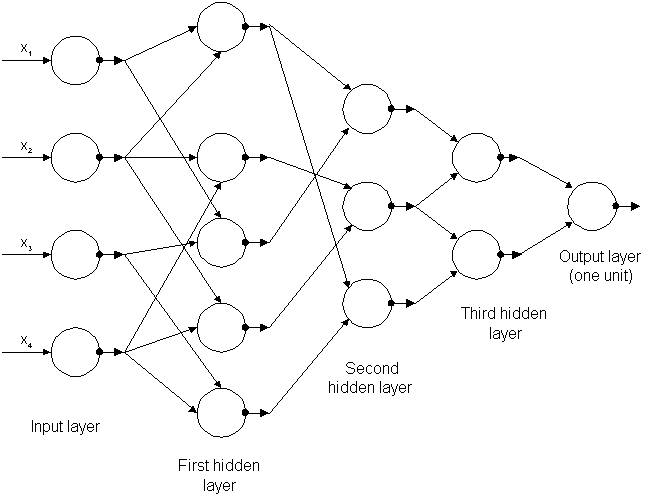}
    \caption[The architecture of the first known deep network by Ivakhnenko]{The architecture of the first known deep network by Ivakhnenko et al.~\cite{ivakhnenko1966cybernetic}.\protect\footnotemark}
    \label{fig:fig1}
\end{figure}

More than half a century ago, Ivakhnenko et al.~\cite{ivakhnenko1966cybernetic} introduced deep MLP (Figure~\ref{fig:fig1}), where thin but deep models (three hidden layers) with polynomial activation functions were used. The authors used statistical methods to select the best features in each layer, and forwarded these features to the next layer until the output layer is reached. Finally, they used layer-by-layer backpropagation algorithm to train the network. A deeper network with eight layers was introduced in~\cite{ivakhnenko1971polynomial} which was trained using the Group Method of Data Handling (GMDH) algorithm.

In 1980, a network with multiple convolutional and pooling layers was introduced in~\cite{fukushima1980neocognitron}, where it was trained using reinforcement learning. The challenge for this model was the training of the multiple layers. At that time, backpropagation of errors was an inefficient and incomplete form to train such deep models.

\footnotetext{\url{https://goo.gl/vcqNjP}}

LeCun gave in 1989 the first efficient and practical application of backpropagation at Bell Labs~\cite{lecun1989backpropagation}. He applied backpropagation to a deep convolutional network in order to classify the handwritten digits of the MNIST dataset. This approach achieved good results. Unfortunately, it consumed a lot of time, which rendered it impractical for many years. In 1993, RNN were introduced to solve the time consumption problem. RNN learned by unsupervised learning, which was implemented and used with very deep learning tasks (more than 1,000 subsequent layers)~\cite{hochreiter1997long}. After that, a DL method called the long short-term memory (LSTM) for RNN was proposed by Hochreiter and Schmidhuber in 1997~\cite{hochreiter1997long}, where it was used in the deep learning tasks that require memories of events (like speech). Moreover, it avoided the vanishing gradient problem at which no learning signals reached to early layers in the network during training the deep network.

A big shift in the field of DL occurred when more people started to use the graphics processing units (GPUs) in the training process. This increased the computational speed allowing NN to produce better results by using more training data. However, training using huge amounts of data brought back to light the vanishing gradient problem. To solve this problem, the models were learning in a layer-by-layer fashion using unsupervised learning. This required the features of early layers to be initialized with suitable features beforehand  (pre-trained). However, in supervised learning, the features at early layers
need to be adjusted during learning process. One solution, which is the pre-training solution, was initially developed for RNN in 1992~\cite{schmidhuber1992learning} and for feedforward networks in 2006~\cite{hinton2006reducing}. A second solution for the vanishing gradient problem in RNN was the LSTM~\cite{hochreiter1997long}.

In the year 2011, the rapid increase in the speed of GPUs reached its glory, which led many researchers such as Ciresan et al.~\cite{ciresan2011flexible} to train deep networks without using pre-training techniques and started to introduce deep learning networks that were constructed from convolutional layers, max-pooling layers, and several fully connected layers followed by a final classification layer~\cite{martinez2013learning}. Krizhevsky et al.~\cite{krizhevsky2012imagenet} used a similar architecture with rectified linear activation functions and dropout function. Since then, the research in DL using GPUs has accelerated rapidly.

\subsection{Convolutional Neural Networks (CNN)}

Convolution, which is widely used in DL networks, is a mathematical process used to mix input data function, $g$, with convolutional kernel (filter), $f$, in order to produce a transformed feature map (a modified version of the original input) as shown in the following equation.
\begin{equation}
(f\times g)(t) = \int_{-\infty}^{\infty}f(\tau)g(t-\tau)d\tau
=\int_{-\infty}^{\infty}f(t-\tau)g(\tau)d\tau
\end{equation}

\begin{figure}
    \centering
    \includegraphics[width=0.5\textwidth]{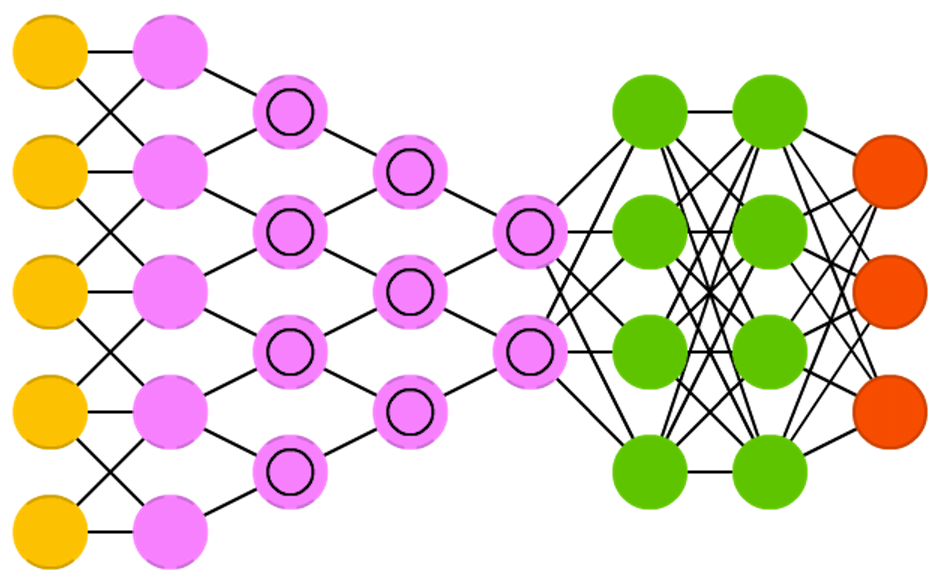}
    \caption[Convolutional Neural Network]{Convolutional Neural Network.\protect\footnotemark}
    \label{fig:fig2}
\end{figure}

Convolution has different applications in many fields such as probability, statistics, computer vision, imaging, etc. In probability theory, convolution is similar to cross-correlation, while in statistics, it updates the weights over normalization of input vector. Figure~\ref{fig:fig2} shows a 
Convolutional Neural Network (CNN or ConvNet), depicted in Figure~\ref{fig:fig2}, is a network with multiple layers of convolutions applying a nonlinear activation function like ReLU or Tanh (mostly used for image processing). CNN layers filter input data to produce useful feature map information. These layers have parameters that are updated repeatedly to produce the desired output. In feedforward NN with fully connected layers, each input neuron is connected to all neurons of the next layer. However, with CNN, after calculation of the output by applying convolutions on the input data, each node only connects itself with the closest neighboring neurons (local connectivity). Convolutional layers sizes shrink as they become deeper.

\footnotetext{\url{http://www.asimovinstitute.org/neural-network-zoo/}}

The architecture of CNN primarily has three types of layers including \textit{convolutional layers}, \textit{pooling layers}, and \textit{fully connected layers}.
The convolutional layers are the main block of a CNN that do most of the computational operations. The pooling layers are applied after the convolutional layers, where these layers partition the input data into non-overlapping sets (windows), and reduce each set to a single value (subsampling) by applying a max operation (outputs the maximum value in each set) to the result of each filter. The pooling layers have benefits of reducing the spatial size of data, reducing the number of parameters, reducing computations, and controlling overfitting. Finally, after all generated features are combined, they are used to find the final classification via fully connected layers, whose neurons are fully connected to all activations in the previous layer.

Many CNN architectures exist such as: 
\begin{itemize}
\item LeNet:
Developed by LeCun et al.~\cite{lecun1998gradient} in the 1990s, LeNet was used to read zip codes, recognize characters, etc.

\item AlexNet: 
Developed by Krizhevsky et al.~\cite{krizhevsky2012imagenet}, AlexNet was used in computer vision tasks. It has a similar architecture to LeNet, but with deeper, bigger, and more stacked convolutional layers on top of each other.

\item GoogleNet:
Introduced by Szegedy et al.~\cite{szegedy2015going} from Google, GoogleNet had a reduced number of parameters in the network (it had 4M parameters, compared to AlexNet's 60M).

\item VGGNet: 
Introduced by Simonyan and Zisserman~\cite{simonyan2014very}, the goal of VGGNet's architecture was to prove that a good performance depends on the depth of the network.

\item ResNet (Residual Network):
Introduced by He et al.~\cite{he2016deep}, ResNet's architecture does not have fully connected layers at the end of the network. Moreover, it uses batch normalization.
\end{itemize}

\subsection{Recurrent Neural Networks (RNN)}
\begin{figure}
    \centering
    \includegraphics[width=0.4\textwidth]{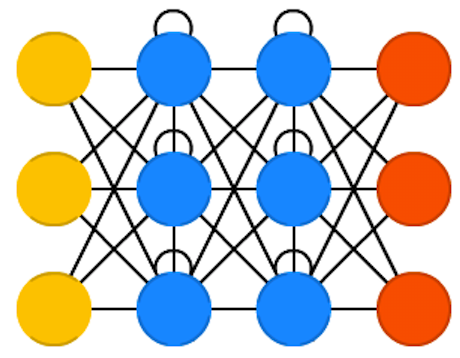}
    \caption[Recurrent Neural Network]{Recurrent Neural Network.\protect\footnotemark[11]}
    \label{fig:fig6}
\end{figure}

RNN is a type of NN, first introduced in 1990 by Elman~\cite{elman1990finding}. After that, Elman and others began to develop the concept of RNN. In 1993, a modified RNN model was developed to solve very deep learning tasks which required more than 100 subsequent layers.
RNN connections between neurons form a directed cycle (fed data from previous layer and from themselves) as shown in Figure~\ref{fig:fig6}. This makes it available to be used for sequential information. Because RNN is built in a way that fits sequential information, it is used in many tasks in NLP, speech recognition, image capturing, language translation, etc.

Unlike Simple NN and CNN models,
in RNN, the output is dependent on the previous computations. Thus, a RNN has an internal memory to save previous computations. For instance, RNN is popular in NLP tasks where it predicts next word depending on previous words in any given sequence using the internal memory in neurons of the hidden layers.

\begin{figure}
    \centering
    \includegraphics[width=0.4\textwidth]{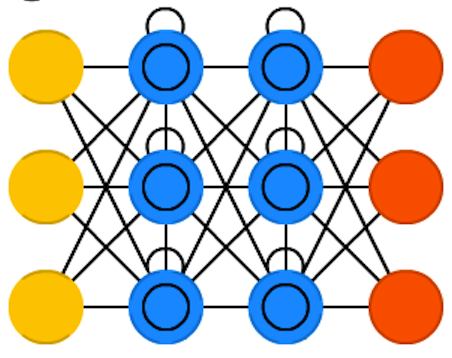}
    \caption[LSTM Neural Network]{LSTM Neural network.\protect\footnotemark[11]}
    \label{fig:fig7}
\end{figure}

Long short-term memory (LSTM) is a special type of RNN (shown in Figure~\ref{fig:fig7}) proposed in 1997 by Hochreiter and Schmidhuber~\cite{hochreiter1997long} in order to solve the vanishing gradient problem. 
It uses LSTM neurons (memory cells with three gates: input, output and forget) instead of simple neurons in the hidden layers. Additionally, LSTM are designed in a way to avoid the long-term dependency problem.
While simple RNN can ``remember'' previous information for short time periods, LSTM can remember them for much longer time periods. However, if the information is not used, it will be lost.

An LSTM cell is more complex than simple (or vanilla RNN) cells as it has a memory to store previous sequences. Each cell contains gates that manage its state and output. Each gate within a unit uses the sigmoid activation function to decide whether it is triggered or not, which makes it conditional to change state or add information.
These three gates within a memory cell are: \textbf{forget gate} to decide what information to discard from each LSTM cell, \textbf{input gate} to decide the update of memory state depending on input values and \textbf{output gate} to decide what to output based on the input and the memory of the LSTM cell.

\section{Deep Learning Frameworks}
\label{sec:frameworks}

The frameworks considered in this comparative study are: CNTK, TensorFlow and Theano. Moreover, we use Keras on top of these frameworks as discussed later. All of these frameworks provide flexible APIs and configuration options for performance optimization.
%
Software versions of the frameworks\footnote{This work was conducted in the summer of 2017. The versions we consider were the latest ones. Since then, these frameworks have been updated.}
are shown in Table~\ref{table:comparative} and their properties are shown in Table~\ref{table:properties}.

\begin{table}
\caption{Frameworks used for this comparative study}\label{table:comparative}
\centering
\begin{tabular}{ccc}
\hline
Framework & Major Version & Github Commit ID\\
\hline
CNTK & 2.0 & 7436a00\\
TensorFlow & 1.2.0 & 49961e5\\
Theano & 0.10.0.dev1 & 8a1af5b\\
Keras & 2.0.5 & 78f26df\\
\hline
\end{tabular}
\end{table}

\begin{table}
\caption{Properties of the considered frameworks}\label{table:properties}
\centering
\begin{tabular}{ccccc}
\hline
Property & CNTK & TensorFlow & Theano & Keras\\
\hline
Core & C++ & C++ & Python & Python\\
CPU & \checkmark & \checkmark & \checkmark & \checkmark\\
Multi-Threaded CPU & \checkmark & Eigen & Blas, conv2D, Limited OpenMP & \checkmark\\
GPU & \checkmark & \checkmark & \checkmark & \checkmark\\
Multi-GPU & \checkmark & \checkmark & X (experimental version) & \checkmark\\
NVIDIA cuDNN & \checkmark & \checkmark & \checkmark & \checkmark\\
\hline
\end{tabular}
\end{table}

\subsection{CNTK}
Microsoft Cognitive Toolkit (CNTK) is an Open source DL framework developed by Microsoft Research~\cite{yu2014introduction} for training and testing many types of NN across multiple GPUs or servers. CNTK supports different DL architectures like Feedforward, Convolutional, Recurrent, LSTM, and Sequence-to-Sequence NN.

In CNTK, a Computational Network learns any function by converting it to a directed graph, where leaf nodes consist of an input values or learning parameters while other nodes represent matrix operation applied to its children. In this case, CNTK has an advantage as it can automatically find the derive gradients for all the computations which are required to learn the parameters. In CNTK, users specify their networks using a configuration file that contains information about the network type, where to find input data, and the way to optimize parameters~\cite{yu2015computational}.

CNTK interface supports different APIs of several languages such as Python, C++ and C\# across both GPU (CUDA) or CPU platforms. According to its developers,\footnote{\url{https://docs.microsoft.com/en-us/cognitive-toolkit/cntk-evaluation-overview}}
CNTK was written in C++ in an efficient way, where it removes duplicated computations in forward and backward passes, uses minimal memory and reduces memory reallocation by reusing them. The framework's installation
is discussed in~\cite{ghadeer_thesis}.

\subsection{Theano}
Theano\footnote{Theano is no longer supported; however, it was so when this paper was written.\\\url{https://groups.google.com/forum/\#!topic/theano-users/7Poq8BZutbY}} is an open source Python library developed at MILA lab at the University of Montreal as a compiler for mathematical expressions that lets users and developers optimize and evaluate their expressions using NumPy's syntax (a Python library that supports a large and multi-dimensional arrays)~\cite{bergstra2010theano,al2016theano}. Theano starts performing computations automatically by optimizing the selection of computations, translates them into other machine learning languages such as C++ or CUDA (for GPU) and then compiles them into Python modules in an efficient way on CPUs or GPUs.

Theano's development started in 2008 and it is more popular on a research and ecosystem platform than many DL libraries. Several software packages have been developed to build on top of Theano, with a higher-level user interface which aims to make Theano easier to express and train different architectures of deep learning models, such as Pylearn2, Lasagne, and Keras. The framework's installation
is discussed in~\cite{ghadeer_thesis}.

\subsection{TensorFlow}
TensorFlow is an open source framework developed by Google Brain Team~\cite{abadi2016tensorflow2}. It uses a single data flow graph, expressing all numerical computations, to achieve excellent performance. TensorFlow constructs large computation graphs where each node represents a mathematical operation, while the edges represent the communication between nodes. This data flow graph executes the communication between sub-computations explicitly, which makes it possible to execute independent computations in parallel or to use multiple devices to execute partition computations~\cite{abadi2016tensorflow2}. The framework's installation
is discussed in~\cite{ghadeer_thesis}.

Programmers of TensorFlow define  large computation graphs from basic operators, then distribute the execution of these graphs across a heterogeneous distributed system (can deploy computation to one or more CPUs or GPUs on a different hardware platforms such as desktops, servers, or even mobile devices). The flexible architecture of TensorFlow allows developers and users to experiment and train a wide variety of deep neural network models, and it is used for deploying machine learning systems into production for different fields including speech recognition, NLP, computer vision, robotics, and computational drug discovery. TensorFlow uses different APIs of several languages such as Python, C++, and Java for constructing and executing a graph (Python API is the most complete and the easiest to use).\footnote{\url{https://www.tensorflow.org/}}
The framework's installation
is discussed in~\cite{ghadeer_thesis}.

\subsection{Keras}

Keras is an open source DL library developed in Python. It runs on top of CNTK, Theano or TensorFlow frameworks. Keras was founded by Google engineer Chollet in 2015 as a part of the research project ONEIROS (Open-ended Neuro-Electronic Intelligent Robot Operating System). Keras is designed in a way that allows fast expression with deep neural networks and easy and fast prototyping (modularity and extensibility)~\cite{chollet2015keras}. The framework's installation
is discussed in~\cite{ghadeer_thesis}.

\section{Methodology}
\label{sec:method}

The goal of this experimental study is to compare the aforementioned frameworks (Theano, TensorFlow and CNTK) by using them to train CNN and RNN models on standard benchmark datasets of classical problems in image processing (MNIST, CIFAR-10, and Self-driving Car) and NLP (Penn TreeBank and IMDB). We then evaluate each framework's performance through the following metrics:
\begin{itemize}
\item Running time
\item Memory consumption
\item CPU and GPU utilization
\item Number of epochs
\end{itemize}

We aim at comparing the aforementioned frameworks using a GPU-equipped laptop that runs Windows 10 operating system, and has the following specifications:
 \begin{itemize}
 \item Intel Core i7-6700HQ CPU @ 2.60GHz (4 cores)
 \item 16 GB RAM
 \item 64-bit operating system, x64-based processor
 \item NVIDIA GEFORCE GTX 960m graphics card (Laptop), with PCI Express 3.0 bus support, equipped with 4 GB GDDR5 memory and 640 CUDA cores.
 \end{itemize}
 
It is worth mentioning that our goal is to compare the resources consumed by each framework to reach a certain accuracy level for each problem. So, we experimented with different epoch counts in order to make sure the accuracy for all frameworks are close to each other.

\subsection{Benchmark Datasets}
In this subsection, we discuss the datasets used in our experiments.

\subsubsection{MNIST}
\begin{figure}
    \centering
    \includegraphics[width=0.4\textwidth]{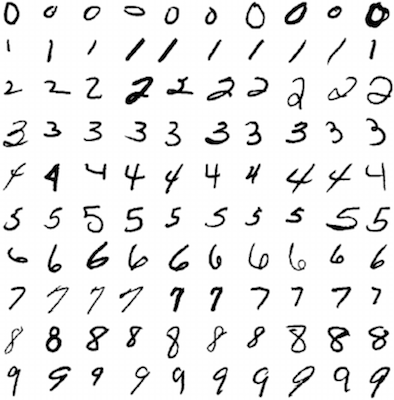}
    \caption[MNIST digits]{MNIST digits.\protect\footnotemark}
    \label{fig:mnist22}
\end{figure}

The MNIST (Mixed National Institute of Standards and Technology) dataset (shown in Figure~\ref{fig:mnist22})
is a computer vision database for handwritten digits. It is widely used for training and testing in the field of machine learning~\cite{lecun1998gradient}.
MNIST has a training set of 60,000 images and a testing set of 10,000 images. It is a subset of a larger set available from NIST. Each image is $28\times~28$ pixels which can be represented as a big array of numbers. We can flatten this array into a vector of $28\times 28= 784$ numbers. Each image in MNIST has a corresponding label, a number between 0 and 9 representing the digit appearing in the image. See Figure~\ref{fig:mnist22}.

\footnotetext{\url{https://goo.gl/Gm8xR7} }

Our goal is to construct a CNN to classify MNIST images. The training will be carried out on both CPU and GPU environments using different frameworks including the aforementioned ones. We then evaluate the performance of each framework.

\subsubsection{CIFAR-10}
The CIFAR-10 dataset is one of the 80 million images datasets, collected by Krizhevsky et al.~\cite{krizhevsky2012imagenet,krizhevsky2009learning}.
It consists of 60,000 $32\times32$ color images evenly distributed over ten classes: airplane, automobile, bird, cat, deer, dog, frog, horse, ship, and truck. There are 50,000 training images and 10,000 test images.

\begin{figure}
    \centering
    \includegraphics[width=0.5\textwidth]{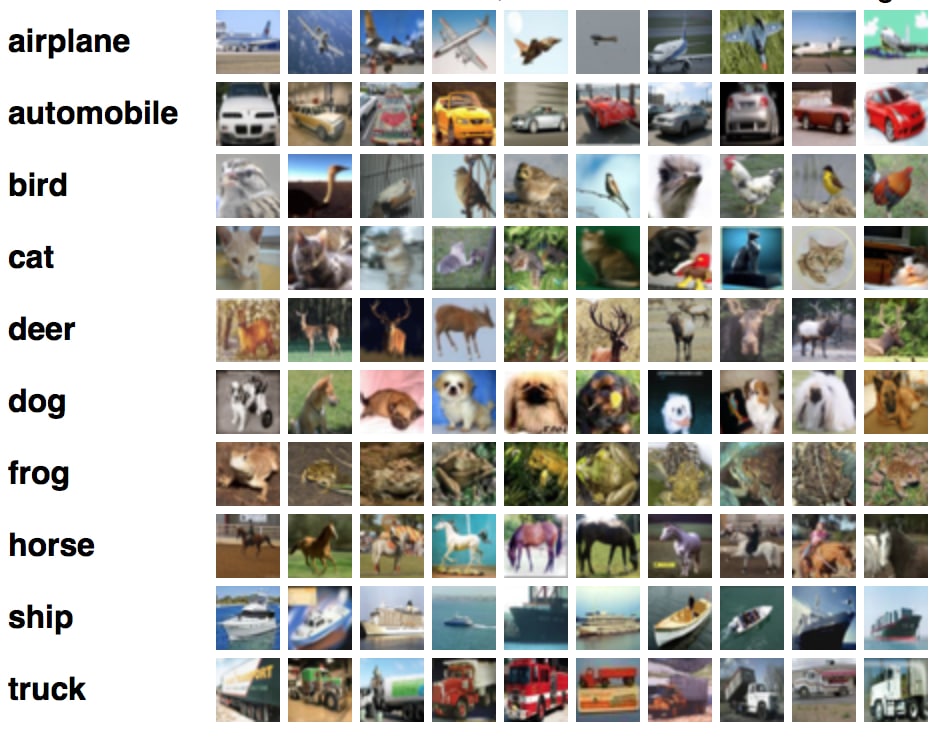}
    \caption[CIFAR-10 dataset classes]{CIFAR-10 dataset classes.\protect\footnotemark}
    \label{fig:cifar10}
\end{figure}

Figure~\ref{fig:cifar10} shows the classes in the dataset, as well as 10 random images from each class. The classes are completely mutually exclusive. I.e., there is no overlap between them. For instance, the ``Automobile'' class includes sedans, SUVs, etc. On the other hand, the ``Truck'' class includes only big trucks. To avoid overlap, neither one of these two classes includes pickup trucks.

\footnotetext{\url{https://www.cs.toronto.edu/~kriz/cifar.html}}

\subsubsection{Penn TreeBank}
\begin{figure}
    \centering
    \includegraphics[width=0.6\textwidth]{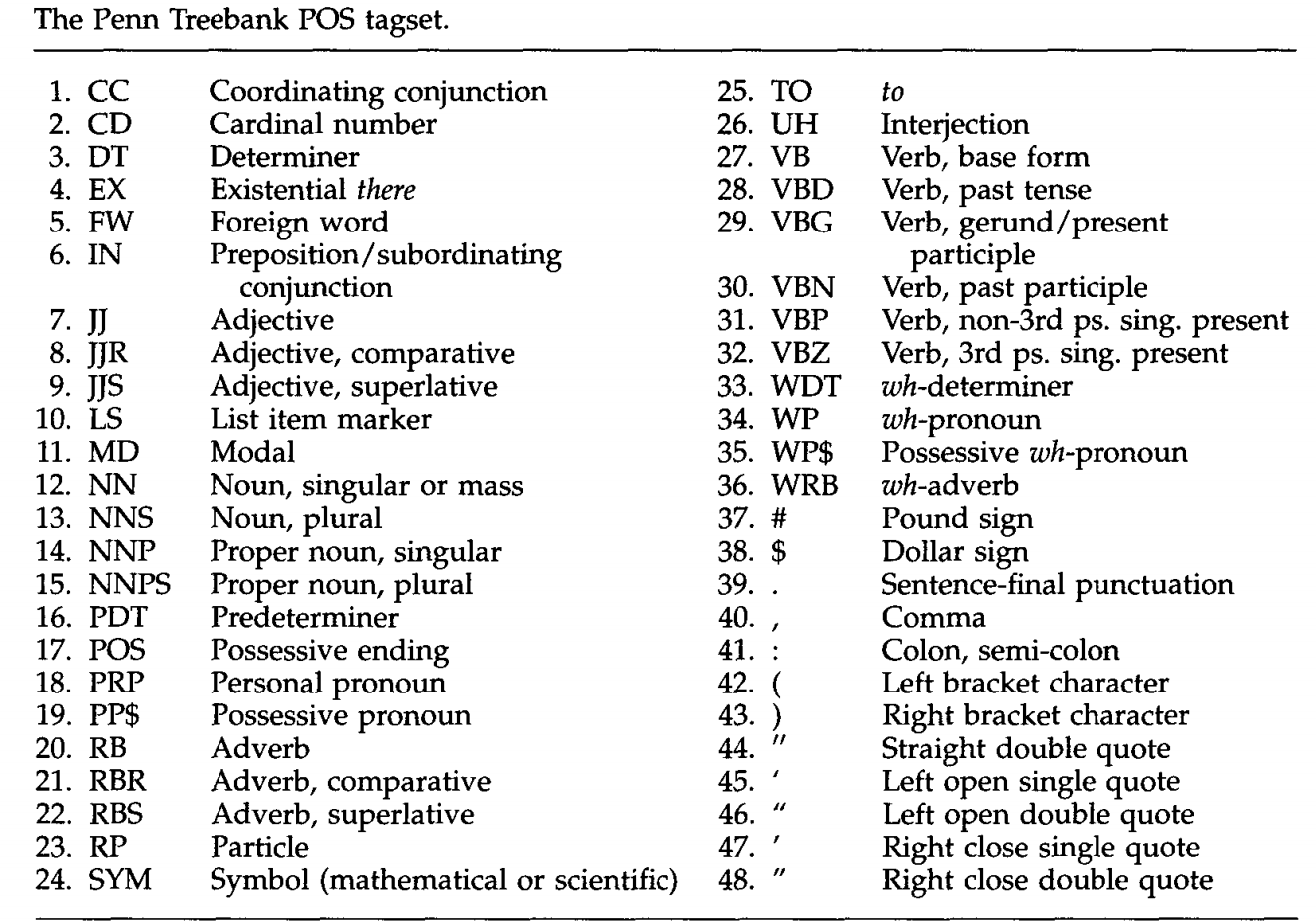}
    \caption[PTB POS tagset]{PTB POS tagset.\protect\footnotemark}
    \label{fig:tagset.png}
\end{figure}

In 1993, Marcus et al.~\cite{marcus1993building} wrote a paper on constructing a large annotated corpus of English called the Penn TreeBank (PTB). They reviewed their experience with constructing one large annotated corpus that consists of over 4.5 million words of American English. The project was divided into phases. For the first three-year phase, the corpus was annotated for part-of-speech (POS) tag information (See Figure~\ref{fig:tagset.png}). Moreover, half of the corpus was annotated for skeletal syntactic structure.

\footnotetext{\url{http://wenhoujx.blogspot.com/2013/06/penn-treebank-pos-tags.html}}

The dataset is large and diverse. It includes the Brown Corpus (retagged) and the Wall Street Journal Corpus, as well as Department of Energy abstracts, Dow Jones Newswire stories, Department of Agriculture bulletins, Library of America texts, MUC-3 messages, IBM Manual sentences, WBUR radio transcripts, and ATIS sentences.

\subsubsection{IMDB}
\begin{figure}
    \centering
    \includegraphics[width=0.7\textwidth]{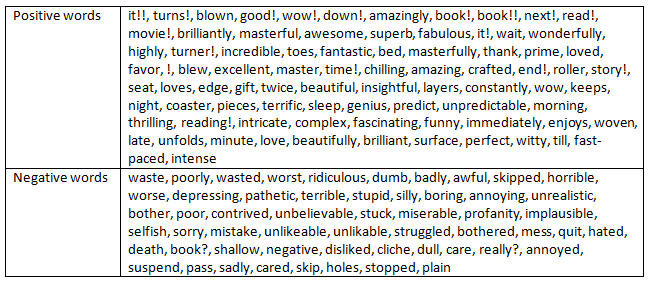}
    \caption[Positive/Negative movie reviews sentiment for IMDB]{Positive/Negative movie reviews sentiment for IMDB.\protect\footnotemark}
    \label{fig:imdb.png}
\end{figure}

The IMDB dataset~\cite{maas2011learning} is another example of applying CNN, which is an online dataset of information regarding films, TV programs and video games. It consists of 25,000 reviews labeled by the sentiment (positive/negative) of each review. The reviews have been preprocessed and encoded as integers in a form of a sequence of word indexes (See Figure~\ref{fig:imdb.png}). Words are indexed by overall frequency in the dataset, so that the index $i$ encodes the $i$th most frequent word in the data in order to allow operations of quick filtering.

\footnotetext{\url{https://goo.gl/S8a6P4}}

\subsubsection{Self-Driving Car}
\begin{figure}
    \centering
    \includegraphics[width=0.7\textwidth]{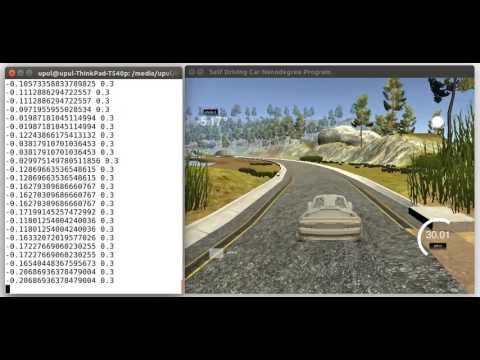}
    \caption[Training Self-Driving Car dataset via Udacity's simulator]{Training Self-Driving Car dataset via Udacity's simulator.\protect\footnotemark}
    \label{fig:self.jpg}
\end{figure}

This dataset uses a Udacity's Self-Driving Car simulator as a testbed for training an autonomous car. This work started in the 1980s with Carnegie Mellon University's Navlab and ALV projects~\cite{wallace1985first}. The training phase starts with activating the simulator which is an executable application. A user initiates the service of collecting the data for training followed by collecting the data as images and saving them locally on the computer. So, the framework can take these images and train them.
The training is done via distinguishing the image's edges which are taken by the three cameras laying on the front of the car in the simulator.
After the training phase is done, the testing phase begins by taking the file generated
whenever the performance in the epoch is better than the previous best.
Finally,the last generated file is executed in order to make the car drive autonomously to observe the testing phase results. See figure~\ref{fig:self.jpg}.

\footnotetext{\url{https://github.com/upul/Behavioral-Cloning}}

\subsection{Networks Architecture}

CNN is used for the MNIST, CIFAR-10, IMDB and Self-Driving Car datasets, where a different network architecture is used for each dataset.
When applying CNN on both Theano and Tensorflow, Keras is used for coding all experiments, while, in CNTK, Keras is used only for the Self-Driving car and IMDB datasets. Thus, MNIST and CIFAR-10 experiments are done without Keras.

For the MNIST and CIFAR-10 datasets, two convolutional layers with ReLU activation function are used after the input layer. The activation function is used to reduce the training time and to prevent vanishing gradients. After each CNN layer, a max-pooling layer is added in order to down-sample the input
and to reduce overfitting. In the max-pooling layer, the stride value must be specified with which the filter is slid. When the stride is one, the filter (window) is moved one pixel at a time. When the stride is two, the filter moved two pixels at a time. This will produce smaller output volumes spatially. After each max-pooling layer, the dropout method is used in order to reduce overfitting by forcing the model to learn many independent representations of the same data through randomly disabling neurons in the learning phase. The sequential architecture (layers part only) used on the MNIST and CIFAR-10 datasets are shown in~\cite{ghadeer_thesis}.

Another example of applying CNN is on the Self-Driving Car dataset, where the network has the same components as the ones used with the MNIST and CIFAR-10 datasets, but with deeper model that consists of five convolutional layers with Exponential Linear Unit (ELU) activation function. The sequential architecture of this CNN is shown in~\cite{ghadeer_thesis}.
The convolutional layers are used for feature engineering. The fully connected layer is used for predicting the steering angle (final output). The dropout avoids overfitting and, finally, the ELU activation function is used to solve the problem of the vanishing gradient.

The final example of applying CNN is on the IMDB dataset. The movie reviews in this dataset are composed of sequences of words of different lengths. These words are encoded by mapping movie reviews to sequences of word embeddings where words are mapped to vectors of real numbers; the network architecture consists of an embedding layer followed by a 1D convolution layer which is used for temporal data followed by a global max-pooling operation. These sequences are padded to have the same size as the largest sequence because they have different lengths. The sequential architecture used on IMDB dataset is shown in~\cite{ghadeer_thesis}.

The other neural network type we consider is RNN with LSTM. One of the most popular uses of LSTM is for text analysis tasks such as the ones associated with the Penn TreeBank (PTB) dataset. Word-level prediction experiments on PTB was adopted, which consists of 929k training words, 73k validation words, and 82k test words. It has 10k words in its vocabulary. We trained models of two sizes (small LSTM and medium LSTM) using the same architecture presented in~\cite{zaremba2014recurrent}.

To evaluate language models of the PTB implementation, a special metric called a perplexity is used, where better prediction accuracy is achieved when perplexity value is as low as possible.
Perplexity is the inverse of probability definition. This means that minimizing perplexity value is the same as maximizing probability.
The goal of applying PTB dataset is to match a probabilistic form which assigns probabilities to sentences. This process is done by predicting next words in a text given a history of previously located words. 
LSTM cells represent the core of the model which processes one word at a time and computes probabilities of the possible values for the next word in the sentence. A vector of zeros is used for the memory state of the network to get initialized and updated after reading each word.

In the small LSTM model, two hidden layers (with 200 LSTM units per layer) are used with Tanh activation function. The weights are initialized to 0.1. We trained it for four epochs with a learning rate of one (number of epochs trained with initial learning rate), and then the learning rate is decreased by a factor of two after each epoch (the decay of the learning rate for each epoch after four epochs), for a total of 13 training epochs. The size of each batch is 20, then the network is unrolled for 20 steps. The sequential architecture used for PTB dataset is shown in~\cite{ghadeer_thesis}.

\section{Results and Discussion}
\label{sec:res}

In this section we discuss the results of our experiments.
Table~\ref{table:proc} shows the CPU and GPU processing times for each dataset.
For the image classification datasets (MNIST and CIFAR-10), one can observe the superiority of CNTK over TensorFlow and Theano in terms of GPU and CPU multithreading; however, in CIFAR-10 using 8, 16 and 32 threads in CPU, TensorFlow was faster than CNTK. On the other hand, Theano revealed to be more time consuming than other frameworks.
Transitioning to sentiment analysis dataset (IMDB), CPU multithreading was not performed because CNTK is written in Python in which multithreading is not supported. Without CPU multithreading (CPU uses the default number of existing physical cores which are equal one thread per core), the superiority of TensorFlow is revealed in both CPU and GPU environments.
The results for
the text analysis dataset (Penn TreeBank)
shows the superiority of TensorFlow over CNTK and Theano, for CPU with 8 threads as well as the case in GPU. Moving forward to video analysis dataset (Self-Driving Car),
the superiority of TensorFlow is revealed in both CPU and GPU environments, while CNTK showed to be more time consuming than the other two frameworks.

\begin{table}
\caption{Processing time for each dataset (measured in seconds)}\label{table:proc}
\centering
\begin{tabular}{ccccccccc}
\hline
Dataset & Environment & Threads & CNTK & TensorFlow & Theano\\
\hline

\multirow {2}{*}{MNIST} & CPU & 1 & 847 & 5130 & 3560 \\
&CPU &2 & 630 & 3180 & 2500 \\ 
&CPU &4 & 574 & 2070 & 2260\\ 
&CPU &8 & 560 & 1740 & 2060\\ 
&CPU &16 & 567 & 1920 & 2050\\ 
&CPU &32 & 588 & 2010 & 2050\\ 
&GPU & -& 66.67 & 328.93 & 377.86 \\
\hline

\multirow{2}{*}{CIFAR-10} & CPU & 1 & 20196 & 25905 & 26700 \\ &CPU &2 & 14520 & 16610 & 18700 \\ &CPU &4 & 13662 & 11550 & 17250\\ &CPU &8 & 11484 & 9955 & 15800\\ &CPU &16 & 11550 & 10340 & 15850\\ &CPU &32 & 11649 & 10835 & 15750\\ &GPU & -& 926 & 2166.4 & 2386.1 \\
\hline

\multirow{2}{*}{IMDB} & CPU & 1 & - & 1244 & 538 \\ &CPU &2 & - & 642 & 412 \\ &CPU &4 & - & 390 & 380\\ &CPU &8 & 486 & 290 & 368\\ &CPU &16 & - & 249 & 368\\ &CPU &32 & - & 302 & 384\\ &GPU & -& 73.1 & 62.4 & 220.41 \\
\hline

\multirow{2}{*}{Self-Driving Car} & CPU & 1 & - & \~ 33.3 hours & \~ 50 hours \\ &CPU &2 & - & \~ 19.8 hours & \~ 44.2 hours \\ &CPU &4 & - &\~ 15 hours & \~ 42.6 hours\\ &CPU &8 & \~ 47.6 hours & \~ 14.1 hours & \~ 43.5 hours\\ &CPU &16 & - & \~ 16.4 hours & \~ 43.5 hours\\ &CPU &32 & - & \~ 16.4 hours & \~ 43.5 hours\\ &GPU  & - & \~8.7 hours & \~6 hours & \~6.8 hours \\
\hline

\multirow{2}{*}{Penn TreeBank} & CPU & 1 & - & 40560 & 27066 \\ &CPU &2 & - & 26819 & 23244 \\ &CPU &4 & - & 18733 & 21541\\ &CPU &8 & 4290 & 16407 & 21450\\ &CPU &16 & - & 16848 & 21476\\ &CPU &32 & - & 18369 & 21541\\ &GPU & - & 2106 & 1342.28 & 1630 \\
\hline

\end{tabular}
\end{table}

\begin{figure}
    \centering
    \includegraphics[width=0.7\textwidth]{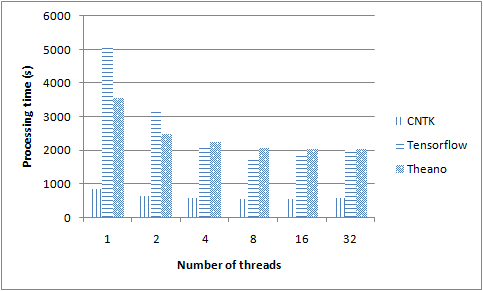}
    \caption{CPU processing time for MNIST dataset}
    \label{fig:mnist}
\end{figure}
\begin{figure}
    \centering
    \includegraphics[width=0.7\textwidth]{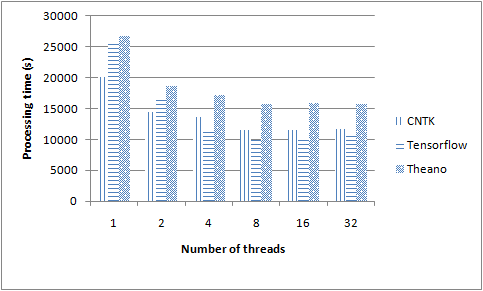}
    \caption{CPU processing time for CIFAR-10 dataset}
    \label{fig:cifar}
\end{figure}
\begin{figure}
    \centering
    \includegraphics[width=0.7\textwidth]{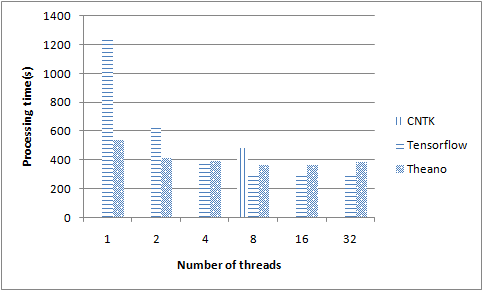}
    \caption{CPU processing time for IMDB dataset }
    \label{fig:imdb2}
\end{figure}
\begin{figure}
    \centering
    \includegraphics[width=0.7\textwidth]{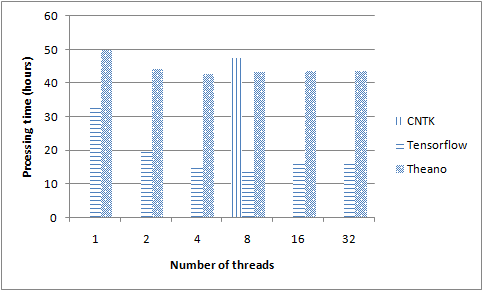}\\
    \caption{CPU processing time for Self-Driving Car dataset }
    \label{fig:car}
\end{figure}
\begin{figure}
    \centering
    \includegraphics[width=0.7\textwidth]{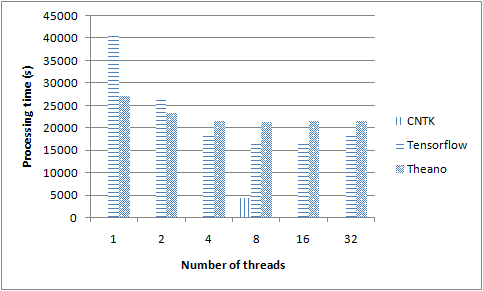}
    \caption{CPU processing time for Penn TreeBnk dataset }
    \label{fig:penn}
\end{figure}
\begin{figure}
    \centering
    \includegraphics[width=0.7\textwidth]{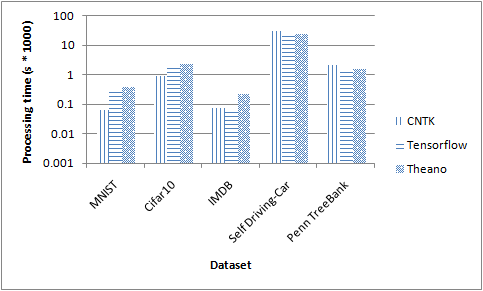}
    \caption{GPU processing time for each dataset }
    \label{fig:gpu}
\end{figure}

Figures (\ref{fig:mnist}\textendash\ref{fig:gpu}) include CPU multithreading and GPU processing time, as well as
Tables~\ref{table:mnisttb}\textendash\ref{table:ptbtb} represent the utilization levels of the CPU, GPU and their memories for each model on each framework.
The processing times clearly show the advantage of GPU over CPU for training deep convolutional and recurrent neural networks. The advantage of fast GPU would be more significant when training complex models with larger data as in the Self-Driving Car dataset.
From the CPU results, the best performance occurred when the number of threads is equal to the number of physical CPU cores, where each thread possesses a single core. In our work we used a laptop with 8 cores, so in each dataset the best performance in terms of processing time was achieved while using 8 threads as shown in figures (\ref{fig:mnist} \textendash~\ref{fig:gpu}).
 The metrics measurement of each framework was conducted to explain the failure of one of the selected frameworks.

We noticed  poor performance of Theano at most datasets comparing to CNTK and TensorFlow.  This could be explained because of low CPU utilization comparing to aforementioned frameworks, meanwhile CNTK and TensorFlow use all available resources (high CPU utilization). CNTK outperformed both TensorFlow and Theano while training MNIST and CIFAR-10 datasets. This achievement is highly likely due to  the use of BrainScript\footnote{\url{https://docs.microsoft.com/en-us/cognitive-toolkit/BrainScript-Network-Builder}}
format which is a custom network description language that makes CNTK more flexible for neural networks customization. On the other hand, 
TensorFlow uses Eigen,\footnote{\url{http://eigen.tuxfamily.org/index.php?title=Main\_Page}}
which is a C++ template library (BLAS library) for linear algebra including matrices, vectors, numerical solvers, and related algorithms. It is used to make TensorFlow perform better than CNTK and Theano in RNN.

Comparing our work to previous work such as Bahrampour’s et al. work presented at ~\cite{bahrampour2015comparative} and Shi et al. work presented at ~\cite{shi2016benchmarking}, we reveal the following findings.
Bahrampour et al.~\cite{bahrampour2015comparative} based their comparative study on three main aspects including speed, hardware utilization, and extensibility. Besides, they used three NN types: CNN, AE, and LSTM to train MNIST, ImageNet, and IMDB datasets on Caffe, Neon, TensorFlow, Theano and Torch frameworks. They
used the following hardware specs; Intel Xeon CPU E5-1650 v2 @3.5GHz (with multi-threading), NVIDIA GeForce GTX Titan X/PCI/SSE2, 32GB DDR3 Memory, and a SSD drive to come
up with the following results. Training on CPU, Torch performed the best followed by Theano while Neon had the worst performance. Moreover, Theano and Torch are the best in terms of extensibility, as well as TensorFlow and Theano were very flexible and Caffe was the easiest to find the performance. Regarding training datasets on GPU, and for larger convolutional and fully connected networks, Torch was the best followed by Neon. For smaller networks Theano was the best. For LSTM, Theano's results were the best, while TensorFlow's performance was not competitive compared with the other studied frameworks.

On the other hand, Shi et al. based their comparative study on two comparative terms including processing time and convergence rate. The neural networks used are fully connected NN, CNN and RNN to train ImageNet, MNIST, and CIFAR10 datasets on Caffe, CNTK, MXNet, TensorFlow, and Torch frameworks. They used the following hardware specs; Two types of multi-threaded CPU: desktop CPU (intel i7-3820) (8 threads) and server-grade CPU (intel xeon E5-2630) (32 threads), as well as three types of NVIDIA GPU (GTX980, GTX1080, Tesla K80), and Two Tesla K80 cards used to evaluate the multi-GPU performance. The results of TensorFlow were the best while using CPU. While using single GPU; on FCN, Caffe, CNTK and Torch performed better than MXNet and TensorFlow. As for small CNN; Caffe and CNTK achieved a good performance, and for RNN (LSTM), CNTK was the fastest (5-10x faster than other frameworks). Using multi-GPU implementation, all frameworks had higher throughput and accelerated the convergence speed. 


In addition to processing time, we also report the utilization levels of the CPU, GPU and their memories for each model on each framework under consideration. These results are shown in Tables~\ref{table:mnisttb}\textendash\ref{table:ptbtb}. The utilization levels for both CPU and GPU are high for all models. The only supersizing numbers are the CPU utilization for Theano, which were very low. The tables also show that the utilization levels are rather small for both types of memory. This applies to all models for all frameworks. However, the tables also show that, in most cases, CNTK had the lowest memory utilization while TensorFlow had the highest. Surprisingly, the case is almost reversed for the video analysis dataset (the Self-Driving Car dataset), where CNTK had the highest utilization and Theano had the lowest. Another unexpected finding of these experiments is that the models of the IMDB generally needed the largest portions of memory.

\begin{table}
\caption{Performance metrics of all models on the MNIST dataset}\label{table:mnisttb}
\centering
\begin{tabular}{cccccccc}
\hline
Metrics & Environment & CNTK & TensorFlow & Theano\\
\hline

\multirow{2}{*}{Accuracy} & CPU & 99.27 & 99.14 & 99.10 \\ &GPU & 99.26 & 99.11 & 99.17 \\

\hline

CPU\% & - & 99.6 & 92.2 & 14.7 \\

\hline
GPU\% & - & 92 & 77 & 95\\

\hline
\multirow{2}{*}{Memory\% } & CPU & 1.7 & 2.2 
& 2.1 \\ &GPU & 3.6 & 5.2 & 4.9 \\

\hline
\multirow{2}{*}{Epochs\# } & CPU & 7 & 15 
& 10 \\ &GPU & 7 & 15 & 10 \\
\hline
\end{tabular}
\end{table}

\begin{table}
\caption{Performance metrics of all models on the the CIFAR-10 dataset}\label{table:cifartb}
\centering
\begin{tabular}{cccccccc}
\hline
Metrics & Environment & CNTK & TensorFlow & Theano\\
\hline

\multirow{2}{*}{Accuracy} & CPU & 82.68 
& 82.26 & 82.29 \\ &GPU & 82.57 & 82.33 & 82.30\\

\hline

CPU\% & - & 99.8 & 87.3 & 15.3 \\

\hline
GPU\% & - & 97 & 73 & 94.5\\

\hline
\multirow{2}{*}{Memory\% } & CPU & 3.2 & 5.3 & 5.1 \\ &GPU & 4.5 & 7.4 & 7.5 
 \\

\hline
\multirow{2}{*}{Epochs\# } & CPU & 33 & 55 
& 50 \\ &GPU & 33 & 55 & 50 \\
\hline
\end{tabular}
\end{table}

\begin{table}
\caption{Performance metrics of all models on the IMDB dataset}\label{table:imdbtb}
\centering
\begin{tabular}{cccccccc}
\hline
Metrics & Environment & CNTK & TensorFlow & Theano\\
\hline

\multirow{2}{*}{Accuracy} & CPU & 88.87 
& 88.68 & 88.72 \\ &GPU & 88.93 & 88.83 & 88.48\\

\hline

CPU\% & - & 94.8 & 92.2 
& 14.6 \\

\hline
GPU\%  & - & 76 & 76 & 88 \\

\hline
\multirow{2}{*}{Memory\% } & CPU & 5.7 & 6.6  & 5.1 \\ &GPU & 6.6 & 9.1 & 7.6 \\

\hline
\multirow{2}{*}{Epochs\# } & CPU & 2 & 3 
& 2 \\ &GPU & 2 & 3 & 2 \\
\hline
\end{tabular}
\end{table}

\begin{table}
\caption{Performance metrics of all models on the Self-Driving Car dataset}\label{table:sdctb}
\centering
\begin{tabular}{cccccccc}
\hline
Metrics & Environment & CNTK & TensorFlow & Theano\\
\hline

\multirow{2}{*}{Accuracy} & CPU & 99.93 
& 99.96 & 99.71 \\ &GPU & 99.97 & 99.97 & 99.73 
\\

\hline

CPU\% & - & 93.2 & 85 & 22 \\

\hline
GPU\% & - & 32.4 & 34 & 31\\

\hline
\multirow{2}{*}{Memory\% } & CPU & 5.3 
& 4.3 & 3.2  \\ &GPU & 6.6 & 6.2 & 5.3 
 \\

\hline
\multirow{2}{*}{Epochs\# } & CPU & 10 & 10 
& 10 \\ &GPU & 10 & 10 & 10 \\
\hline
\end{tabular}
\end{table}

\begin{table}
\caption{Performance metrics of all models on the Penn TreeBank dataset}\label{table:ptbtb}
\centering
\begin{tabular}{ccccccccc}
\hline
Metrics & Environment & CNTK & TensorFlow & Theano\\
\hline

\multirow{2}{*}{Perplexity} & CPU & 113.7 
& 114.79 & 114.57  \\ &GPU & 113.2 & 113.21 & 113.3 \\

\hline

CPU\% & - & 94 & 91.8 & 18.3  \\

\hline
GPU\% & - & 76.6 & 77 & 81\\

\hline
\multirow{2}{*}{Memory\% } & CPU & 1.3
& 2.3 & 4.1 \\ &GPU & 2.2 & 4.5 & 5.4 \\

\hline
\multirow{2}{*}{Epochs\# } & CPU & 13 & 13 
& 13 \\ &GPU & 13 & 13 & 13 \\
\hline

\end{tabular}
\end{table}

\section{Conclusions and Future Work}
\label{sec:conc}

\subsection{Conclusion}
In this paper, we have provided a qualitative and quantitative comparison between three of the most popular and most comprehensive DL frameworks (namely Microsoft's CNTK, Google's TensorFlow and University of Montreal's Theano). The main goal of this work was to help
end users make an informed decision about the best DL framework that suits their needs and resources. To ensure that our study is as comprehensive as possible, we have used multiple benchmark datasets namely MNIST, CIFAR-10, Self-Driving Car, and IMDB which were trained via multilayer CNN network architecture and Penn TreeBank dataset which was trained via RNN architecture. We have run our experiments on a laptop with windows 10 operating system. We have measured performance and utilization of CPU multithreading, GPU and memory.
For most of our experiments, we find out that CNTK's implementations are superior to the other ones under consideration.

\bibliographystyle{unsrt}
\bibliography{references}

\end{document}